\documentclass{IOS-Book-Article}

\usepackage{mathptmx}
\usepackage{graphicx}
\usepackage{amsmath,amssymb}
\usepackage{amsfonts}
\usepackage{soul}\setuldepth{article}
 
\def\hb{\hbox to 10.7 cm{}}

\begin{document}

\pagestyle{headings}
\def\thepage{}

\begin{frontmatter}              

\title{Intuitive Contrasting Map for Antonym Embeddings}

\markboth{}{April 2021\hb}

\author[A]{\fnms{Igor } \snm{SAMENKO}},
\author[B]{\fnms{Alexey} \snm{TIKHONOV}}
and
\author[C]{\fnms{Ivan P.} \snm{YAMSHCHIKOV}}\thanks{Corresponding Author: LEYA Lab, Yandex and Higher School of Economics in St. Petersburg; Kantemirovskaya st. 3; E-mail:
ivan@yamshchikov.info}

\runningauthor{I. Samenko et al.}
\address[A]{Institute of Computational Technologies, Russian Academy of Sciences, Novosibirsk, Russia }
\address[B]{Yandex, Berlin, Germany}
\address[C]{LEYA Lab, Yandex and Higher School of Economics in St. Petersburg, Russia}

\begin{abstract}
This paper shows that, modern word embeddings contain information that distinguishes synonyms and antonyms despite small cosine similarities between corresponding vectors. This information is encoded in the geometry of the embeddings and could be extracted with a straight-forward and intuitive manifold learning procedure or a {\em contrasting map}. Such a map is trained on a small labeled subset of the data and can produce new embeddings that explicitly highlight specific semantic attributes of the word. The new embeddings produced by the map are shown to improve the performance on downstream tasks.
\end{abstract}

\end{frontmatter}
\markboth{April 2021\hb}{April 2021\hb}

\section{Introduction}

Modern word embeddings, such as \cite{mikolov2013efficient}, \cite{pennington2014glove} or \cite{bojanowski2017enriching} are based on the distributional hypothesis \cite{harris1954distributional}. If two words are often used in a similar context, they should have a small cosine similarity between the embeddings. Naturally, such methods often fail to recognize antonyms since antonymous words, e.g., "fast" and "slow", occur in similar contexts. Many researchers address this issue from different angles.

Some authors deal with representations of antonyms, injecting additional information, and improving training procedures. For example, \cite{bian2014knowledge} combine deep learning with various types of semantic knowledge to produce new word embeddings that show better performance on a word similarity task. \cite{ono2015word} combine information from thesauri with distributional information from large-scale unlabelled text data and obtain word embeddings that could distinguish antonyms. \cite{liu2015learning} represent semantic knowledge extracted from thesauri as many ordinal ranking inequalities and formulate the learning of semantic word embeddings as a constrained optimization problem. \cite{kim2016adjusting} develop these ideas further and adjust word vectors using the semantic intensity information alongside with thesauri. \cite{dou2018improving} also use thesauri along with the sentiment to build new embeddings that contrast antonyms. \cite{nguyen2016integrating} improve the weights of feature vectors with a special method based on local mutual information and propose an extension of the skip-gram model that integrates the new vector representations into the objective function. \cite{hill2014not} and \cite{hill2014embedding} show that translation-based embeddings perform better in applications that require concepts to be organized according to similarity and better capture their true ontologic status. \cite{lu2015deep} use these ideas and demonstrate that adding a multilingual context when learning embeddings allows improving their quality via deep canonical correlation analysis. 

Other researchers try to develop novel approaches that are not heavily relying on the distributional hypothesis. For example, \cite{schwartz2015symmetric} introduce word-level vector representation based on symmetric patterns and report that such representations allow controlling the model judgment of antonym pairs. \cite{chen2015revisiting} develop special {\em  contrasting embedding framework}. \cite{nguyen2017distinguishing} train a neural network model that exploits lexico-syntactic patterns from syntactic parse trees to distinguish antonyms. 

All works mentioned above were based on the assumption that antonym-distinguishing information is not captured by modern word embeddings. However, this assumption is frequently questioned in the last several years. \cite{vulic2018specialising} and \cite{vulic2018injecting} show one can inject information on hyponyms, hyperonyms, synonyms, and antonyms to distinguish the obtained embeddings using additional linguistic constraints, see \cite{mrkvsic2017semantic}. Moreover, \cite{etcheverry2019unraveling} come up with a two-phase training of a siamese network that transforms initial embeddings into the ones that clearly distinguish antonyms. \cite{ali2019antonym} develop an architecture of a distiller that extracts information on antonyms out of the pre-trained vectors.

In this work, we demonstrate that Word2Vec \cite{mikolov2013efficient}, GloVe \cite{pennington2014glove}, and especially FastText \cite{bojanowski2017enriching} embeddings contain information that allows distinguishing antonyms to certain extent. This information is encoded in the geometry of the obtained vector space. We propose a very simple and straightforward approach for the extraction of this information. Similarly to \cite{etcheverry2019unraveling} it is based on a siamese network, yet does not require a two-phase training and is more intuitive than the one proposed in \cite{ali2019antonym}. We also show that this approach could be used further to extract other semantic aspects of words out of the obtained embedding space with ease.

The contribution of this paper is as follows: 
\begin{itemize}
\item we demonstrate that modern word embeddings contain information that allows distinguishing synonyms and antonyms; 
\item we show that this information could be retrieved by learning a nonlinear manifold via supervision provided by a small labeled sub-sample of synonyms and antonyms;
\item we demonstrate that concatenation of these new embeddings with original embeddings improves the performance on the downstream tasks that are sensitive to synonym-antonym distinction.
\end{itemize}

\section{Data}
\label{sec:dt}

\begin{figure}[h!]
    \centering

    \includegraphics[scale=0.4]{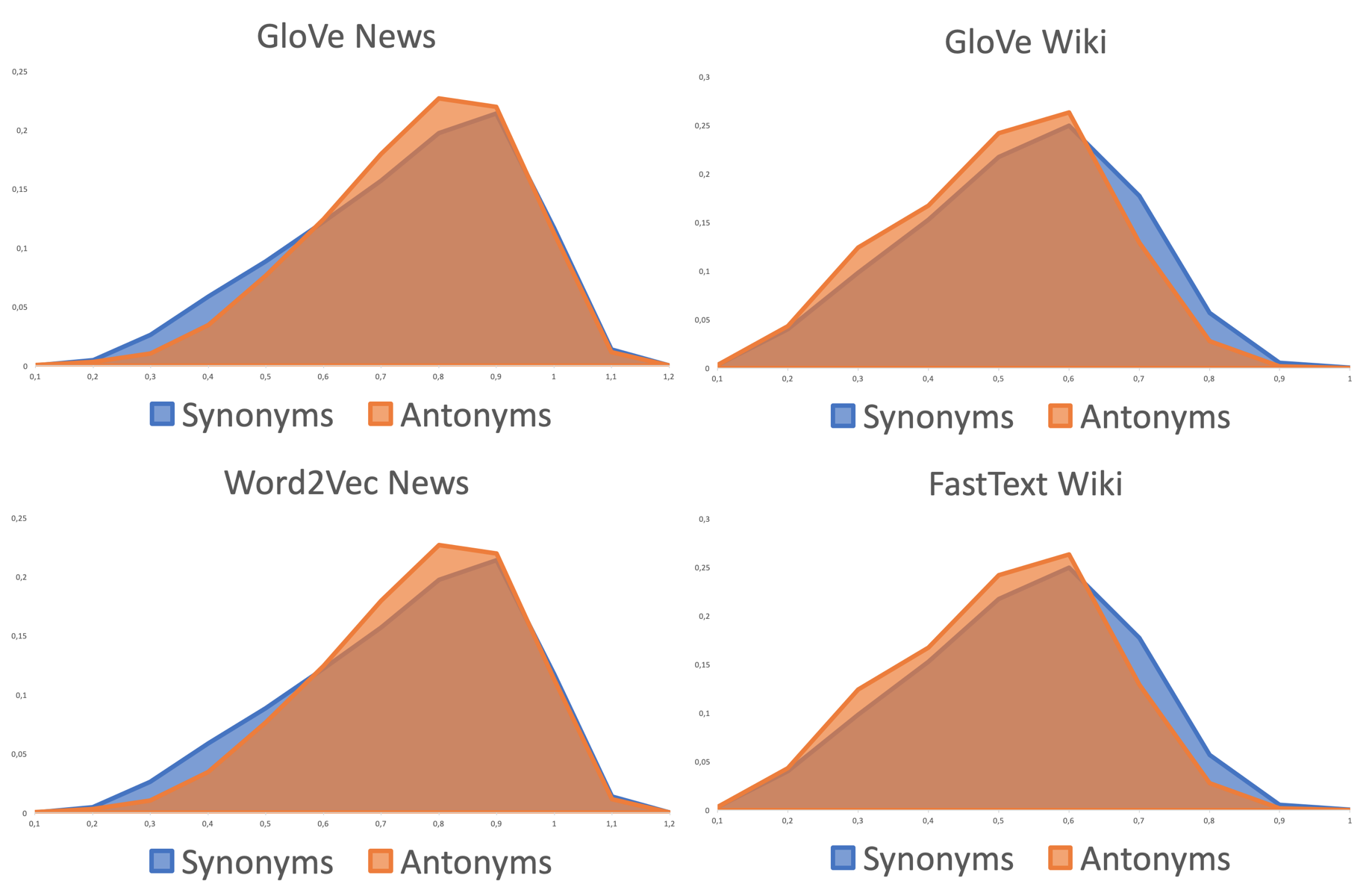}
    \vspace*{0.7cm}
    \caption{Distribution of cosine distances between synonyms and antonyms across four different sets of embeddings.}
    \label{fig:data}
\end{figure}

For the experiments, we used the small supervised set of synonyms and antonyms of English language provided by WordNet\footnote{\scriptsize{https://wordnet.princeton.edu/}} that we enriched with additional data from \cite{nguyen2017distinguishing} and several other publicly available sources\footnote{\scriptsize{https://github.com/ec2604/Antonym-Detection}}.  We tested the methodology described below across multiple modern word embeddings, namely, FastText\footnote{\scriptsize{https://fasttext.cc/}}, GloVe pre-trained on Wikipedia\footnote{\scriptsize{https://nlp.stanford.edu/projects/glove/}} and GloVe pre-trained on Google News\footnote{\scriptsize{https://www.kaggle.com/pkugoodspeed/nlpword2vecembeddingspretrained}} alongside with Word2Vec pre-trained on Google News\footnote{\scriptsize{https://github.com/mmihaltz/word2vec-GoogleNews-vectors}}. In Figure \ref{fig:data} one could see initial distributions of cosines between synonyms and antonyms in four different training datasets respectively.

The WordNet dataset of synonyms and antonyms consists of 99833 word pairs. Synonymic relations are neither commutative nor transitive. For example, "economical" could be labeled as a synonym to "cheap," yet the opposite is not true\footnote{\scriptsize{https://www.thesaurus.com/browse/cheap}}. At the same time, if "neat" is denoted as a synonym to "cool" and "cool" is denoted as a synonym to "serene," this does not imply that "neat" and "serene" are synonyms as well. All data sources used in this paper are in the public domain. To facilitate reproducibility, we share the code of the experiments\footnote{\scriptsize{https://github.com/i-samenko/Triplet-net/}}

We propose the following train-test split procedure that guarantees that the words from the training dataset do not infiltrate the test set. We add pairs to train and test with relative frequencies of 3 to 1. If one of the words in the pair was already in the train or test, we were adding the new pair to the corresponding subset. If one word in the pair occurs both in train and in the test, we deleted such a pair. After such a test-train split, we obtained 80 080 pairs. 65 292 pairs of 26 264 unique words formed the training dataset, and 14 788 pairs of 8737 unique words comprised the test dataset.

Figure \ref{fig:data} seems to back up the widespread intuition that modern embedding can not distinguish synonyms and antonyms. However, in the next sections, this paper demonstrates that this statement does not hold.  

\section{Learning Contrasting Map}

 If one assumes that information allowing to distinguish synonyms and antonyms is already present in the raw embeddings, one could try to extract it by building a manifold learning procedure that would take original embedding as input and try to map it in a new space of representations, where the synonym-antonym contrast becomes explicit. 

The initial embedding space is $\mathbb{R}^m$ with a distance $D_m$ defined on it, and for every word '$w$', for any of its synonyms '$s$', and for any of its antonyms '$a$' the following holds $D_m(w,s) \simeq D_m(w,a)$. A new embedding space of lower dimension $\mathbb{R}^k$ has a corresponding distance $D_k$. One would like to find a map $f: \mathbb{R}^m \rightarrow \mathbb{R}^k$ such that the following holds $D_k(w,s) < D_k(w,a)$ in a new $\mathbb{R}^k$ embedding space. 

\begin{equation} \label{eq:con}
f = \begin{cases}
f: \mathbb{R}^m \rightarrow \mathbb{R}^k,~~~ m>>k; \\
D_k(w,s) < D_k(w,a),~~~ \forall w, s, a.
\end{cases}
\end{equation}

Since the amount of synonyms and antonyms in any given language is growing excessively with the growth of the training sample of texts, one can not check these conditions for every word pair explicitly. One can only use a labeled subset of the vocabulary, where synonyms and antonyms are contrasted already, so it is hard to establish a procedure that would guarantee Inequalities \ref{eq:con}, hence we use $\lesssim$ for the conditions. At the same time, despite the limited size of the training dataset, one would hope that the obtained representations are general enough to distinguish the synonyms and antonyms that are not included in the training data. 

\begin{figure}[h]
    \includegraphics[scale=0.3]{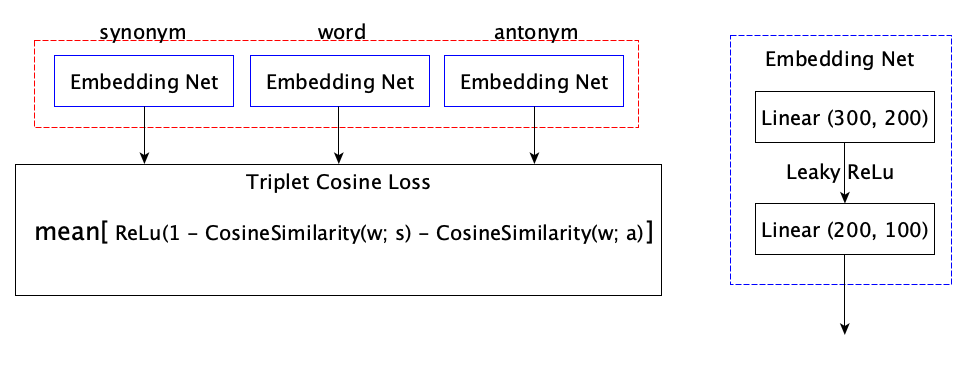}
        \vspace*{0.7cm}
    \caption{Siamese Triplet Network trained to distinguish synonyms and antonyms. EmbeddingNet is the contrasting map $f: \mathbb{R}^m \rightarrow \mathbb{R}^k$. The weights of three EmbeddingNets are shared in the end-to-end training. The resulting architecture is trained to minimize cosine similarities between synonyms and maximize the cosine similarities between antonyms in the transformed low-dimensional embeddings space $\mathbb{R}^k$.}
    \label{fig:cos}
\end{figure}

To train such a map let us regard an architecture, shown in Figure \ref{fig:cos}. It is a 'Siamese' network \cite{bromley1994signature} where weights are shared across three identical EmbeddingNets. Each EmbeddingNet maps the word '$w$', its synonym '$s$' and its antonym '$a$' respectively. The resulting cosine similarities between synonyms and antonyms are simply included in the loss function in such a way that $D_k(w,s)$ is minimized and $D_k(w,a)$ is maximized explicitly. The whole system is trained end-to-end on 65 292 pairs of synonyms and antonyms described in Section \ref{sec:dt}.

\section{Experiments}

\begin{figure}[h!]
    \centering
    \includegraphics[scale=0.4]{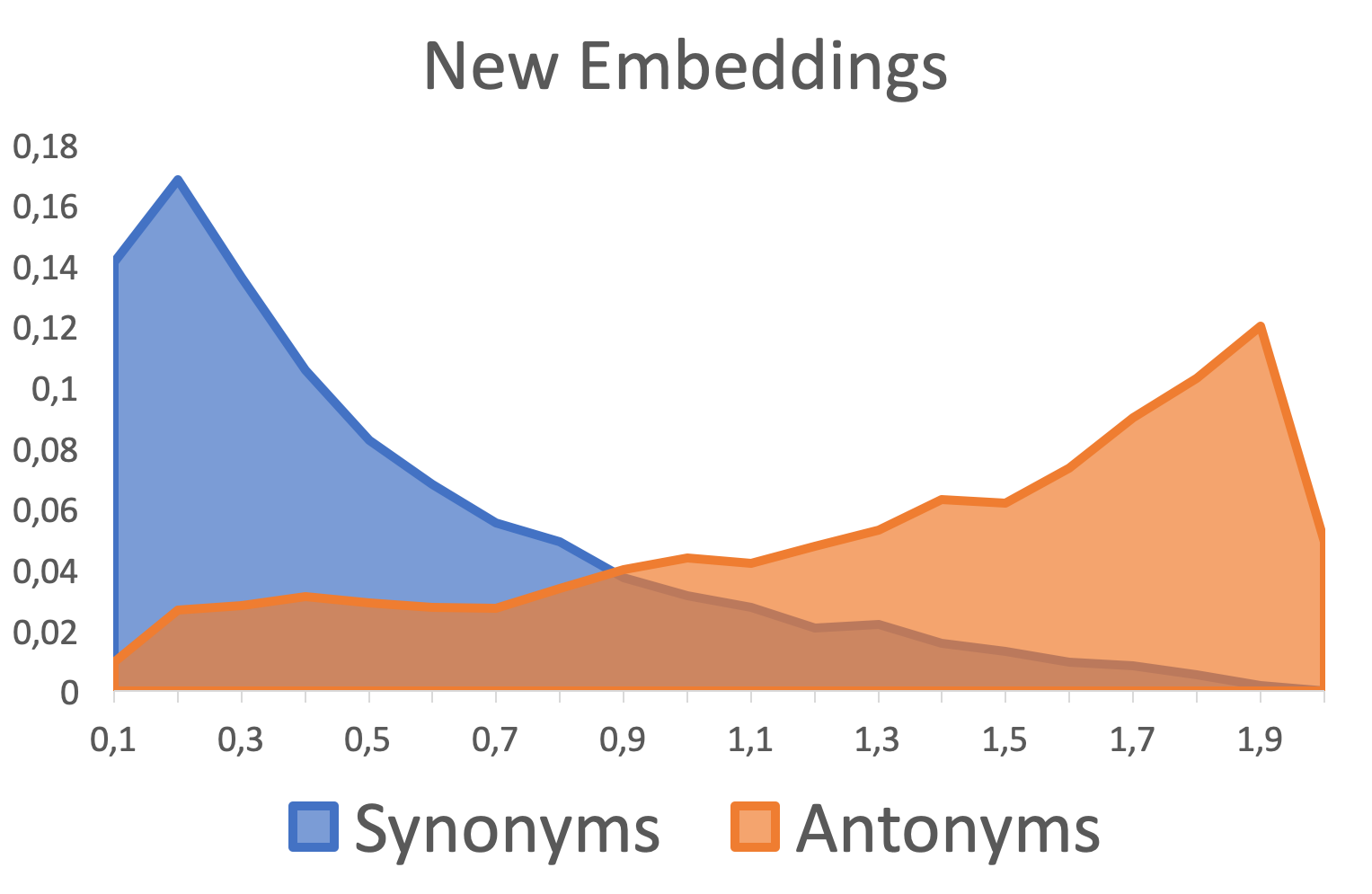}
        \vspace*{0.9cm}
    \caption{Distribution of cosine distances between synonyms and antonyms in the transformed space $\mathbb{R}^k$ for FastText. Test set. Different datasets produce similar results. Distances between synonyms tend to become smaller, distances between antonyms tend to increase.}
    \label{fig:trans}
\end{figure}

 First of all, let us check if the condition listed in Equation \ref{eq:con} is satisfied in the transformed embedding space $\mathbb{R}^k$. Figure \ref{fig:trans} illustrates the distributions of the cosine distances between synonyms and antonyms in $\mathbb{R}^k$ for English FastText embeddings. The situation is drastically improved in contrast with raw embeddings shown in Figure \ref{fig:data}.

One can have a close look at the tails of the distributions shown in Figure \ref{fig:trans}. To simplify further experiments and improve reproducibility we also publish the resulting distances for the test set\footnote{\scriptsize{https://github.com/i-samenko/Triplet-net/}}.

Here are some examples of word pairs that were marked as antonyms in the test dataset, yet are mapped close to each other by the contrast map: \verb"sonic"    — \verb"supersonic"; \verb"fore" —    \verb"aft"; \verb"actinomorphic" — \verb"zygomorphic"; \verb"cable"    — \verb"hawser"; \verb"receive" — \verb"give"; \verb"ceiling" — \verb"floor". Here are some examples of word pairs that were marked as synonyms in the test dataset, yet are mapped far of each other by the contrast map: \verb"financial" —    \verb"fiscal"; \verb"mother" —    \verb"father"; \verb"easy" —    \verb"promiscuous"; \verb"empowered"    — \verb"sceptred"; \verb"formative" — \verb"plastic"; \verb"frank"    — \verb"wiener"; \verb"viii" —    \verb"eighter"; \verb"wakefulness" —    \verb"sleeplessness". One can see that some of the contrasting map errors are due to the debatable labeling of the test dataset, others occur with the words that are rare.

\begin{figure}[h!]
    \centering
    \includegraphics[scale=0.4]{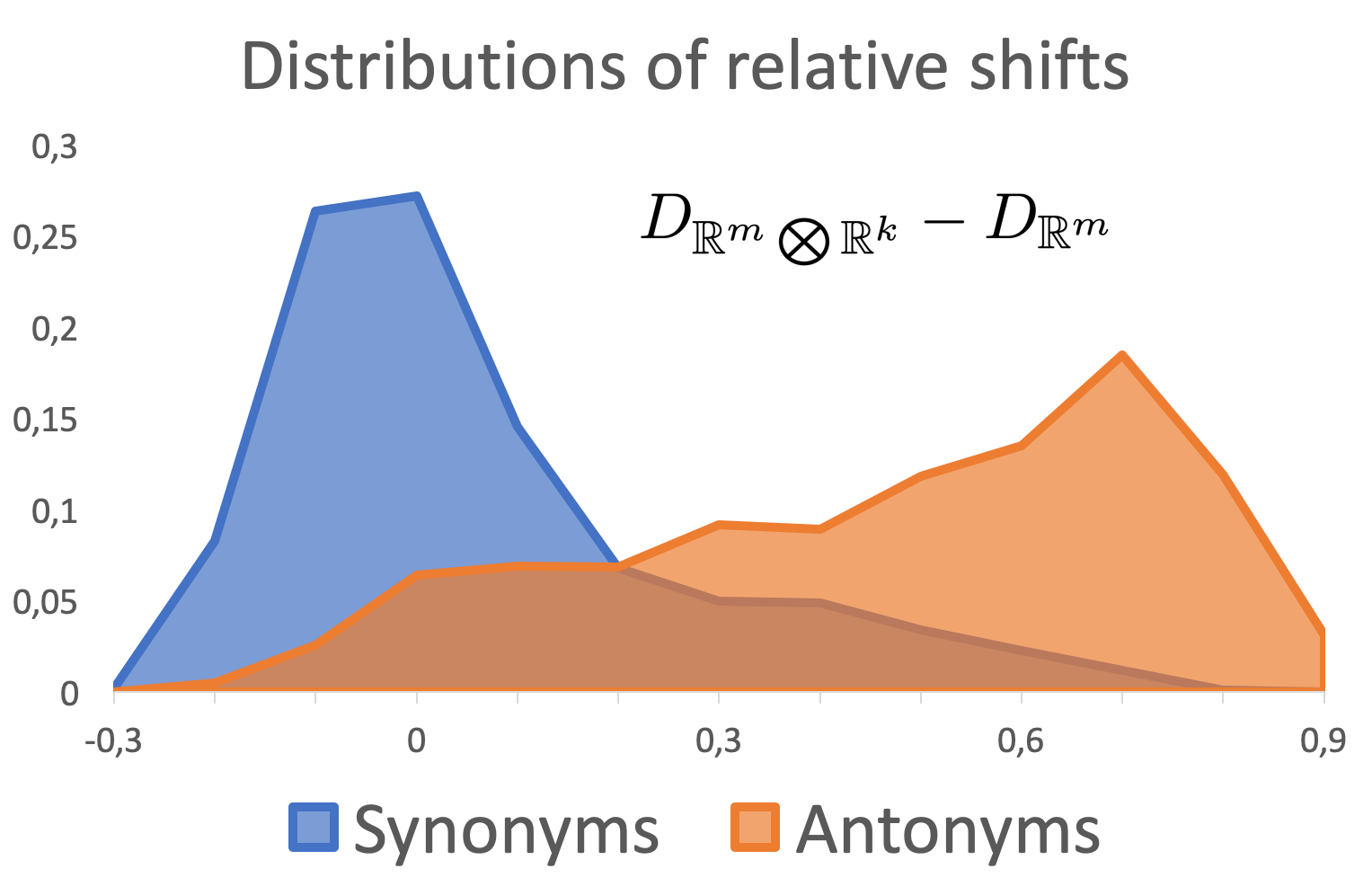}
        \vspace*{0.7cm}
    \caption{Cosine distances between synonyms and antonyms in the raw embeddings and in the space where they are concatenated with the new ones. FastText embeddings. Test set.}
    \label{fig:shifts2}
\end{figure}

To be sure that other properties of the original embeddings are preserved we concatenate new embedding with the old, raw ones. Figure \ref{fig:shifts2} depicts the difference of the pairwise distance between synonyms and antonyms in the space of concatenated embeddings $D_{\mathbb{R}^m \bigoplus \mathbb{R}^k}$ and in the space of raw embeddings $D_{\mathbb{R}^m}$. The distributions are obtained for the test subset of data. The map did not see these word pairs in training. 

We train an XGBoost classifier on four different raw embeddings and check the resulting accuracy of the classifiers on the test subset of synonym and antonym pairs. Table \ref{tab:experimentsemb} clearly shows that the accuracy of a classifier trained on raw embeddings is consistently lower than the accuracy of the same classifier trained on the newly transformed embeddings, produced by the EmbeddingNet. One can also see that a classifier trained on the concatenation of the raw embeddings with the new ones also outperforms the classifier trained solely on the original embeddings. 

\begin{table}
\centering
\small{\begin{tabular}{lllll}
 \hline
Embeddings type &Raw & New & Concatenated \\
\hline
Word2Vec & 0.67 & 0.85 & 0.81 \\
\hline
GloVe Wiki & 0.65 & 0.75 & 0.72\\
\hline
GloVe Google News & 0.67 & 0.84 & 0.78\\
\hline
FastText & 0.73 & {\bf 0.88} & {\bf 0.85}\\
\hline
\end{tabular}}
\caption{Comparison of four different embeddings. For every type of embedding, XGBoost classifier is trained to distinguish two input vectors as synonyms or antonyms.}
  \label{tab:experimentsemb}
\end{table}

FastText embeddings are capturing more than 80\% of synonym-antonym relations with the proposed contrasting map, and more than 70\% out of these relations are captured out of the box. GloVe embeddings seem to contain the least information on the synonym-antonym relations. Further experiments are conducted on FastText embeddings since they capture the most out of synonym-antonym relations. 

To illustrate the potential usage of such embeddings obtained with a contrasting map we run a series of experiments with various NLP datasets that intuitively might need to contrast synonyms and antonyms for the successful performance: binary sentiment classifier for IMDB reviews\footnote{\scriptsize{https://ai.stanford.edu/\~amaas/data/sentiment/}},  binary sentiment classifier for Cornell movie reviews\footnote{\scriptsize{http://www.cs.cornell.edu/people/pabo/movie-review-data/}}, binary classifier to identify toxic comments\footnote{\scriptsize{https://www.kaggle.com/c/jigsaw-toxic-comment-classification-challenge}}, sentiment classifiers across several thematic domains of  Multi-Domain Sentiment Dataset\footnote{\scriptsize{http://www.cs.jhu.edu/\~mdredze/datasets/sentiment/}}.

For every dataset, we trained a logistic regression using pre-trained FastText embeddings and measured its accuracy on the test. Then we retrained the same logistic regression with new concatenated embeddings. Table \ref{tab:down} demonstrates how the usage of the transformed embeddings improves the accuracy on various datasets.

\begin{table}
\centering
\small{\begin{tabular}{llll}
 \hline
Dataset & FastText only & Concatenated\\
\hline
IMDB reviews & 0.86 & 0.88 {\bf (+2.2\%)} \\
\hline
Cornell reviews & 0.75 & 0.76 {\bf (+1.0\%)} \\
\hline
Toxic Comments & 0.94 & 0.95 {\bf (+0.6\%)} \\
\hline
MDSD books & 0.69 & 0.77 {\bf (+11.3\%)} \\

MDSD DVDs & 0.70 & 0.76 {\bf (+8.0\%)} \\

MDSD electronic & 0.72 & 0.78 {\bf (+9.4\%)} \\

MDSD kitchen & 0.78 & 0.80 {\bf (+3.4\%)} \\

MDSD all categories & 0.76 & 0.79 {\bf (+3.6\%)} \\
\hline
\end{tabular}}
\caption{Concatenation of the original FastText embeddings with transformed embeddings improves the accuracy of logistic regression-based classifiers trained on various datasets.}
  \label{tab:down}
\end{table}

\section{Discussion}

The proposed methodology demonstrates that contrary to common intuition modern word embeddings contain information that allows distinguishing synonyms and antonyms. The approach could possibly be scaled to other semantic aspects of the words. In its most general form, the approach allows mapping original embeddings into spaces of lower dimensions that could explicitly highlight certain semantic aspects using a labeled dataset of a limited size. This semantic information can be effectively incorporated into the downstream tasks.

Conceptually, the proposed methodology allows for revisiting the questions of language acquisition in the context of the distributional hypothesis. If one assumes that semantic information attached to a given word is not a rigid structure but depends on the training corpus, it seems that modern embeddings capture these diverse semantic fields successfully, provided the corpus is large enough. This result does not mean that such semantic aspects are explicit and could be immediately extracted out of the embeddings. The spaces of modern word embeddings could be profoundly nonlinear concerning a given semantic attribute of the word. A deeper understanding of the geometric properties of these spaces could significantly improve the quality of the resulting models. Indeed, the very assumption that semantic similarity could be captured with cosine distance in Euclidian space is debatable.

Though the choice of the embedding space and the notion of distance on it both need further, more in-depth investigations, this paper demonstrates the simple methods of representational learning applied to the raw embeddings can distill this implicitly encoded information reasonably well.

\section{Conclusion}

This paper demonstrates that contrary to a widely spread opinion modern word embeddings contain information that allows distinguishing synonyms from antonyms.  This information is encoded in the geometry of the embeddings and could be extracted with manifold learning. The paper proposes a simple and intuitive approach that allows obtaining a {\em contrasting map}. Such a map could be trained on a small subset of the vocabulary and is shown to highlight relevant semantic information in the resulting vector embedding. The new embeddings, in which the information on synonyms and antonyms is disentangled, improve the performance on the downstream tasks. The proposed methodology of contrasting maps could potentially be further extended to other semantic aspects of the words.

\bibliographystyle{vancouver}
\bibliography{bibl}

\end{document}